\documentclass[runningheads]{llncs}

\usepackage[T1]{fontenc}
\usepackage{booktabs} 
\usepackage{times}  
\usepackage{helvet}  
\usepackage{courier}  
\usepackage{url}  
\usepackage{graphicx}  
\usepackage{comment}
\usepackage{amsmath}
\usepackage{multirow}
\usepackage{makecell}
\usepackage{color}
\usepackage{cancel}
\usepackage{hyperref} 
\usepackage{cite}
\usepackage{amsmath,amssymb,amsfonts}
\usepackage{textcomp}
\usepackage{xcolor}
\usepackage{subfig}
\usepackage{marvosym}
\usepackage[ruled,lined,linesnumbered]{algorithm2e}
\def\BibTeX{{\rm B\kern-.05em{\sc i\kern-.025em b}\kern-.08em
    T\kern-.1667em\lower.7ex\hbox{E}\kern-.125emX}}

\newcommand{\method}{CroDiNo-KD}
\newcommand{\CE}{CE}

\newcommand{\greenArrow}{ (\textcolor{green}{$\uparrow$}) }
\newcommand{\redArrow}{ (\textcolor{red}{$\downarrow$}) }

\begin{document}
\title{Revisiting Cross-Modal Knowledge Distillation: \\ A Disentanglement Approach for \\ RGBD Semantic Segmentation}
\titlerunning{\method}

\author{Roger Ferrod\inst{1}(\Letter) \and
Cássio F. Dantas\inst{2,3} \and
Luigi Di Caro\inst{1} \and
Dino Ienco\inst{2,3}}
\authorrunning{R. Ferrod et al.}
\institute{University of Turin, Turin, Italy \\
\email{(roger.ferrod,luigi.dicaro)@unito.it} \and
INRAE, UMR TETIS, Univ. Montpellier, Montpellier, France \and
EVERGREEN, Univ. Montpellier, Inria, Montpellier, France \\
\email{(cassio.fraga-dantas,dino.ienco)@inrae.fr}
}
\maketitle

\begin{abstract}
Multi-modal RGB and Depth (RGBD) data are predominant in many domains such as robotics, autonomous driving and remote sensing. The combination of these multi-modal data enhances environmental perception by providing 3D spatial context, which is absent in standard RGB images. Although RGBD multi-modal data can be available to train computer vision models, accessing all sensor modalities during the inference stage may be infeasible due to sensor failures or resource constraints, leading to a mismatch between data modalities available during training and inference. Traditional Cross-Modal Knowledge Distillation (CMKD) frameworks, developed to address this task, are typically based on a teacher/student paradigm, where a multi-modal teacher distills knowledge into a single-modality student model. However, these approaches face challenges in teacher architecture choices and distillation process selection, thus limiting their adoption in real-world scenarios. To overcome these issues, we introduce \method{} (Cross-Modal Disentanglement: a New Outlook on Knowledge Distillation), a novel cross-modal knowledge distillation framework for RGBD semantic segmentation. Our approach simultaneously learns single-modality RGB and Depth models by exploiting disentanglement representation, contrastive learning and decoupled data augmentation with the aim to structure the internal manifolds of neural network models through interaction and collaboration. We evaluated \method{} on three RGBD datasets across diverse domains, considering recent CMKD frameworks as competitors. Our findings illustrate the quality of \method, and they suggest reconsidering the conventional teacher/student paradigm to distill information from multi-modal data to single-modality neural networks. Source code is available \href{https://github.com/rogerferrod/CroDiNo-KD}{here}.


\end{abstract}

\section{Introduction}
\label{sec:intro}
Multi-modal information, such as RGB and Depth (RGBD) imagery, is becoming predominant in a plethora of diverse domains including robotics, autonomous driving, augmented reality, healthcare and remote sensing. The combination of these complementary sources of information significantly enhances environmental perception by enriching traditional 2D images with 3D spatial context provided by the Depth modality.

Despite the advantages of multi-modal learning, real-world deployment faces practical challenges. While multi-modal data may be available during training, operational constraints often limit modality availability at inference time due to sensor failures or budget restrictions. This can result in a mismatch between training and testing data, which can impede the practical deployment of an RGBD multi-modal model. To address this challenge, it is essential to design frameworks that are resilient to missing modalities at test time, transferring multi-modal knowledge available during training into single-modality models that operate solely on either RGB or Depth information at inference time. To this purpose, Cross-Modal Knowledge Distillation (CMKD) frameworks have been introduced~\cite{Xue2022TheMF}. Conversely to traditional knowledge distillation techniques, which typically transfers knowledge from a large model to a smaller one using the same input data~\cite{Bucila2006ModelC}, CMKD enables the transfer of information across modalities. Existing CMKD frameworks typically adopt a teacher/student paradigm, transferring knowledge from a multi-modal teacher to a single-modality student. However, these methods are sensitive to design choices such as teacher architecture, fusion mechanisms and knowledge distillation techniques. Moreover, they require substantial computational resources associated with the training of multiple neural network models: a multi-modal teacher and separate single-modality students, one for each target modality.


With the aim to advance cross-modal knowledge distillation for RGB and Depth imagery, we introduce \method{} (Cross-Modal Disentanglement: a New Outlook on Knowledge Distillation), a novel framework that goes beyond conventional teacher/student paradigm, dominant in the CMKD field. Rather than relying on a multi-modal teacher model to guide single-modality RGB or Depth models,  \method{} relaxes the need for a teacher model through a collaborative training strategy where single-modality models interact with each other via carefully designed loss functions. Our approach removes design decisions related to the teacher architecture and fusion mechanism and teacher/student knowledge distillation techniques. Furthermore, \method{} reduces training resources in terms of computational time and parameter size while achieving superior results to recent approaches based on the common teacher/student paradigm. 

Specifically, \method{} jointly trains two single-modality neural networks using disentangled representation and contrastive learning. This process structures each model's internal manifold into modality-invariant and modality-specific features, capturing both shared and unique information from RGB and Depth modalities. Finally, the training process enables a flexible data augmentation strategy, eliminating the constraints of conventional CMKD framework that require paired augmentation techniques between modalities.

In summary, our contributions are threefold: 
\begin{itemize}
\item[(i)] We introduce a novel framework for cross-modal knowledge distillation based solely on the joint training of two single-modality models, offering an alternative to the traditional multi-modal teacher/student paradigm; 
\item[(ii)] We are the first to explore disentanglement representation learning jointly with contrastive learning for RGBD cross-modal knowledge distillation, demonstrating the benefits of structuring internal models manifold into modality-invariant and modality-specific information;
\item[(iii)] We provide insights and discussion on the advantages of our framework beyond classification results, analyzing resource efficiency in terms of both computational training time and model size (parameters count).
\end{itemize}

We validate the effectiveness of \method{} on three RGBD benchmarks for semantic segmentation across different application domains, demonstrating superior performance compared to recent state-of-the-art methods especially designed for semantic segmentation under cross-modal knowledge distillation.

\section{Related Work}
\label{sec:related}
Knowledge Distillation (KD) is the process of transferring information from a large model (teacher) to a smaller one (student). Originally envisioned in \cite{Bucila2006ModelC} for classification tasks with the aim to provide a compact, smaller and faster  model, yet performing comparably to the wider teacher model, it has been further refined and formalized by \cite{Hinton2015DistillingTK}, where KD has been commonly implemented via a Kullback-Leibler (KL) divergence between teacher and student predictions. The KD framework can be formulated as follows:
\begin{equation}
\label{eq:kd}
    \mathcal{L} = \alpha\mathcal{L}_{task} + (1-\alpha)\mathcal{L}_{KD}
\end{equation}
where $\mathcal{L}_{task}$ is the task-specific loss and $\mathcal{L}_{KD}$ the KL divergence between student and teacher predictions. By changing the way the KD loss is used, one could distill different kinds of knowledge: response-based \cite{Jin2023MultiLevelLD}, feature-based \cite{Guo2023ClassAT} or relation-based \cite{Huang2022KnowledgeDF}.

Beyond traditional approaches, KD has also been successfully applied to multi-modal learning~\cite{Xue2022TheMF}. Taking inspiration from the the standard KD process, one can distill the knowledge from a multi-modal teacher to single-modality students~\cite{Liu2023LiteMKDAM}, or from a single-modality teacher to a student working on a different modality~\cite{Hafner2018CrossmodalDF}. 
Considering semantic segmentation, cross-modal KD has been proven to be effective over different applications~\cite{minhao}. For example, studies such as~\cite{acnet, Xu2023FRNetFN, Hazirbas2016FuseNetID, Couprie2013IndoorSS, Yang2023PixelDC, Lee2017RDFNetRM, Jiang2018RedNetRE} explored RGBD segmentation with standard KD frameworks, while~\cite{Lv2024ContextAwareIN, 9108585} performed similar experiments on RGBT (RGB+Thermal) dataset. Following works extended the standard cross-modal knowledge distillation approach by adding a generative task~\cite{masked}, via prototype learning~\cite{prototype} or by decomposing the KD loss function into magnitude and angular terms~\cite{liu}.

Differently from standard learning processes, disentanglement representation learning aims to explicitly decompose the feature representation into semantic factors carrying explainable and meaningful information \cite{wang2024disentangled}. Leveraged also in multi-modal scenarios (e.g., \cite{Tsai2018LearningFM, 9718029}) it can be used to learn modality-specific and modality-invariant features for the downstream task \cite{Xu2021PredictPA, 10.1145/3503161.3547935, 9879072}. In particular, in \cite{bmvc} the authors successfully exploited disentanglement ---together with adversarial learning--- for cross-modal knowledge distillation in the context of scene classification. Inspired by this pioneering work, we further extended this research path onto dense classification, more precisely semantic segmentation.

\section{Method}
\label{sec:method}
With the objective to overcome the limitations of current teacher/student paradigm, here we introduce \method, a new cross-modal knowledge distillation framework that combines disentanglement representation learning, contrastive learning and decoupled data augmentation. Our approach simultaneously trains two single-modality models -- one for RGB and another for Depth imagery -- by exploiting modality interaction and collaboration during the training stage.

\subsection{Proposed framework}
\label{sec:arch}
\begin{figure*}[t]
    \centering
    \includegraphics[width=1\textwidth]{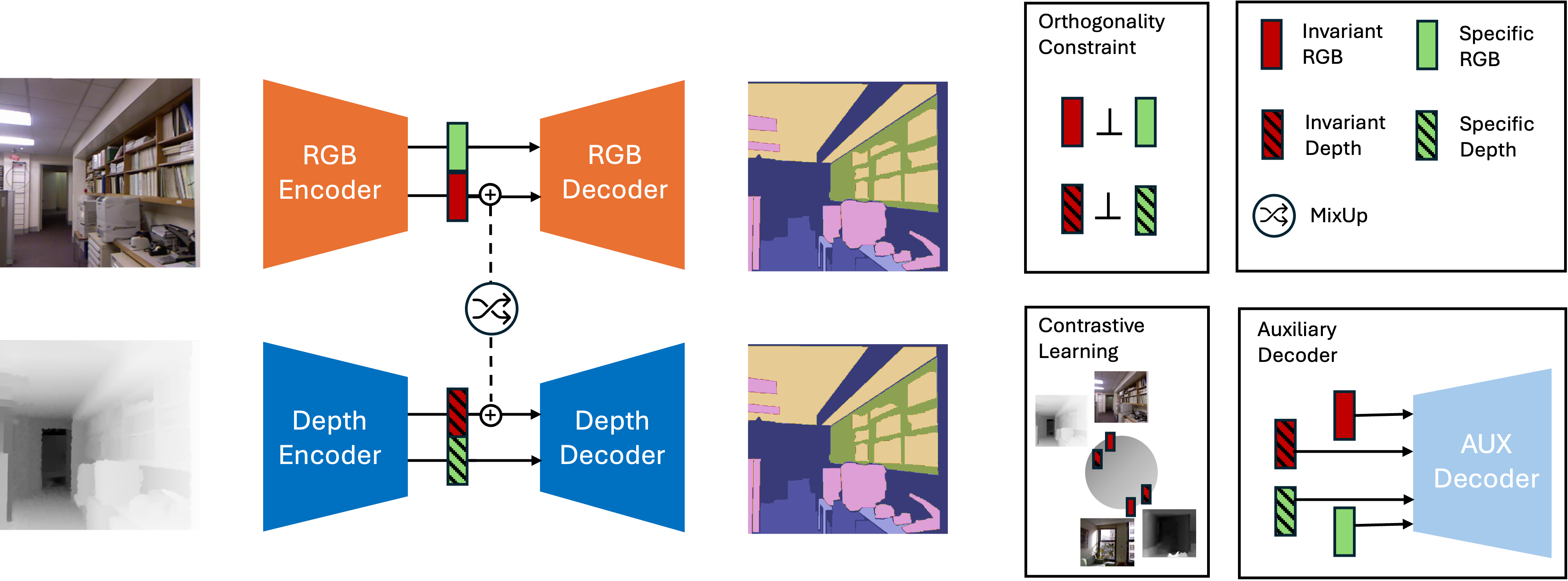}
    \caption{Overview of the \method{} architecture, composed by two encoder-decoder models, for both RGB and Depth modalities. In addition an auxiliary decoder and a set of loss functions are adopted to enforce the desired disentanglement properties between modalities, i.e., modality-invariant and modality-specific features for both RGB and Depth information.\label{fig:model}}
\end{figure*}

The overall framework, depicted in Figure~\ref{fig:model}, consists of: i) two separate encoder-decoder models and ii) an auxiliary decoder, all trained with a set of carefully designed loss functions to structure the internal manifold representation of the single-modality models into modality-invariant and modality-specific features.

Given a batch of RGB images $X_{RGB}$ and the corresponding Depth images $X_{D}$, with $X_{RGB}  \in \mathbb{R}^{B \times H \times W \times 3}$ and $X_{D}  \in \mathbb{R}^{B \times H \times W \times 1}$, we first encode them, via convolutional neural networks, into embedding representations $Z_{RGB}$ and $Z_{D}$, respectively. Denoting generically $Z_{m} \in \mathbb{R}^{B \times h \times w \times F}$ with $m \in \{RGB, D\}$ we have:
\begin{equation}
   Z_{m} = Enc_m(X_m)
\end{equation}
where $B$ is the batch size, $H \times W$ the spatial dimension of the RGB and Depth images, $h \times w$ the spatial dimension of the embedding representations and $F$ the number of output channels.

For each modality, once the encoded representation $Z_m$ is obtained, we divide it into two separate embeddings $Z_m^{inv}$ and $Z_m^{spc}$, with $Z_m^{inv}, Z_m^{spc} \in \mathbb{R}^{B \times h \times w \times F/2}$. During training, we then encourage $Z_m^{inv}$ (resp. $Z_m^{spc}$) to encode modality-invariant (resp. modality-specific) information. 

To generate segmentation outputs, the decoder takes as input the concatenated representation $[Z_m^{inv} : Z_m^{spc}]$, where $[:]$ denotes concatenation along the feature dimension.
The auxiliary decoder, used only during training, follows a similar architecture but takes only half the channel dimension as input. While the decoder included in the main model relies on $[Z_m^{inv} : Z_m^{spc}]$ as input, the auxiliary one works separately on $Z_{RGB}^{inv}$, $Z_{RGB}^{spc}$, $Z_{D}^{inv}$ and $Z_{D}^{spc}$ to enforce every individual embedding representation to encode relevant information for the segmentation task.

With the aim to encourage invariant representation across modalities, 
we introduce a feature mixup strategy~\cite{mixup}. Precisely, we blended the RGB and Depth invariant embeddings, following the equations below, with $\lambda \in [0,1]$: 
\begin{align}
    &\tilde{Z}_{RGB}^{inv} = \lambda Z_{D}^{inv} + (1-\lambda)Z_{RGB}^{inv} \nonumber \\ 
    &\tilde{Z}_{D}^{inv} = \lambda Z_{RGB}^{inv} + (1-\lambda)Z_{D}^{inv}
    \label{eq:mixup}
\end{align}
The augmented images $\tilde{Z}_m^{inv}$ are then processed by the main decoder and contribute to the final loss computation together with the original ones. 

\paragraph{\textbf{Losses:}} To enhance the performance of single-modality models through mutual interaction and collaboration, we design a set of loss functions that shape the models' internal manifold. The first term is a task-specific segmentation loss, modeled through Cross Entropy. More formally, we have:
\begin{equation}
    \mathcal{L}^m_{seg} =  \CE \left(Dec_m([Z_{m}^{inv}:Z_{m}^{spc}]), Y \right) 
\label{eq:seg}
\end{equation}
where 
$\CE$ denotes the pixel-wise cross-entropy loss, $Y\in \{1, \dots, C\}^{B \times H \times W}$ is the ground-truth segmentation map over $C$ classes and $Dec_m$ the decoder for the modality $m \in \{RGB, D\}$.

Then, to explicitly constrain embeddings to encode complementary information (i.e., modality-invariant and modality-specific) we enforced orthogonality between the modality-invariant and modality-specific embeddings of the same modality ${m \in \{RGB,D\}}$ as follows:

\begin{equation}
    \mathcal{L}^m_{\perp} = 
    \frac{1}{B}
    \sum_{b=1}^B
    \sum_{i =1}^{h}
    \sum_{j =1}^{w}
    sim( Z_{m}^{inv}[b,i,j,:], Z_{m}^{spc}[b,i,j,:])
\label{eq:perp}
\end{equation}
where $Z_{m}[b,i,j,:] \in \mathbb{R}^{F/2}$ is the feature vector at spatial location $ij$ in the feature map corresponding to the $b$-th sample in the batch 
and $sim(u, v) = \frac{ u \cdot v }{||u||_2 ||v||_2}$  denotes the cosine similarity between vectors $u$ and $v$.

Furthermore, we introduce a contrastive term to bring $Z_{RGB}^{inv}$ and $Z_{D}^{inv}$ closer together, to force the representation to be invariant with respect to the modality. To this end, we relied on the InfoNCE loss~\cite{infonce} with a negative Euclidean distance, contrasting a positive example (i.e., an RGB-Depth pair of the same instance) with in-batch negatives (i.e., all the remaining invariant embeddings inside the batch, both RGB and Depth). Let $p_{m}^{(b)} \in \mathbb{R}^{F/2}$ be the L2 normalized feature vector of the $b$-th instance obtained via spatial average pooling from $Z_{m}^{inv}$, that is:
\begin{equation}
    p_{m}^{(b)} = \frac{\rho_{m}^{(b)}} {\|\rho_{m}^{(b)}\|_2}, \quad \text{with } 
    \rho_{m}^{(b)} = \frac{1}{hw} \sum_{i =1}^h \sum_{j=1}^{w}  {Z_{m}^{inv}}[b,i,j,:],
    \label{eq:pooling}
\end{equation}
the contrastive loss is then formulated as:
\begin{align}
    &\mathcal{L}^{RGB}_{con} = - \frac{1}{B} \sum_{i=1}^B log \frac{exp(-\| p_{RGB}^{(i)} - p_{D}^{(i)} \|_2/\tau)}{  \sum_{m \in \{RGB, D\}}  \sum_{j \neq i} {exp(-\|p_{RGB}^{(i)}  - p_{m}^{(j)} \|_2/\tau)}}  \nonumber \\
    &\mathcal{L}^D_{con} = - \frac{1}{B} \sum_{i=1}^B log \frac{exp(-\| p_{D}^{(i)} - p_{RGB}^{(i)} \|_2/\tau)}{  \sum_{m \in \{RGB, D\}}  \sum_{j \neq i} {exp(-\|p_{D}^{(i)}  - p_{m}^{(j)} \|_2/\tau)}}  
\label{eq:con}
\end{align}
where $\tau$ is the temperature parameter. In the first equation, the RGB modality serves as the anchor, while in the second equation, the Depth modality takes this role. In both cases, the numerator represents the positive pair, which corresponds to the embeddings of the same instance across different modalities. The denominator contains the negative samples, comprising all other invariant embeddings from both modalities within the batch, excluding the anchor itself.

To ensure that the embeddings independently encode relevant information for the segmentation task, we added an auxiliary segmentation loss that processes each embedding individually:
\begin{equation}
    \mathcal{L}^m_{aux} =
    \CE \left(Dec_{Aux}(Z_{m}^{inv}), Y \right) + \CE \left(Dec_{Aux}(Z_{m}^{spc}), Y \right)
\label{eq:aux}
\end{equation}
with $Dec_{Aux}$ the auxiliary decoder.

Finally, the loss optimized by \method{} is an unweighted combination of all the loss terms previously introduced:

\begin{equation}
    \mathcal{L}_{tot} =  \sum_{m \in \{RGB,D\}} \mathcal{L}^m_{con}  + \mathcal{L}^m_{seg} + \mathcal{L}^m_{\perp} + \mathcal{L}^m_{aux}
\label{eq:loss}
\end{equation}

\paragraph{\textbf{Training procedure:}}
The training process, outlined in Algorithm \ref{alg:training}, runs over a predefined number of epochs ($N_{ep}$). For each batch in an epoch, it starts by augmenting the RGB and Depth images, as commonly done in prior works~\cite{acnet, sigma}.
However, unlike conventional CMKD frameworks, which enforce paired transformations for both RGB and Depth images, our approach relaxes this constraint by allowing independent per-modality augmentations, a strategy we term as \textit{decoupled augmentation} (lines 3–4). Since RGB and Depth losses are computed separately, this strategy enables greater augmentation flexibility compared to approaches based on the standard teacher/student paradigm, where augmentation consistency across modalities is required.


Next, we extract both domain-invariant and domain-specific embeddings for RGB and Depth images using their respective encoders (lines 5–6). To enhance domain-invariant representations, we leverage feature mixup (line 9), which blends RGB and Depth features, enriching the decoder’s training samples (lines 10–12). Additionally, an auxiliary decoder is used to enforce task discrimination for both modality-invariant and modality-specific feature representations, independently (lines 13–15).

To accommodate disentanglement representation learning properties, we use an orthogonality constraint between domain-invariant and domain-specific embeddings (line 16) and we rely on a contrastive loss (line 17) to encourage the RGB and Depth representations of the same instance to be closer to each other while ensuring separation from other instances within the same batch, regardless of the modality. Finally, the total loss is computed as an unweighted sum of all the previously computed losses across RGB and Depth modalities (line 19), back propagating the signal and updating the framework components accordingly.

\begin{algorithm}
\caption{\label{alg:training} \method \, training procedure}
         \SetKwInOut{Input}{input}
        \Input{RGB-Depth labeled dataset $\mathcal{D} = \{(X_{RGB}, X_D, Y)^{(i)}\}_{i=1}^N$}
    \SetKwBlock{Beginn}{beginn}{ende}
    \For{\textnormal{epoch} $\in \{1, \dots, N_{ep}\}$}{
        \ForAll{ \textnormal{batches} $(X_{RGB}, X_{D}, Y) \in \mathcal{D}$}{
            
            \tcp{Decoupled Augmentations}
            $X_{RGB}$ = Aug($X_{RGB}$); 
            
            $X_{D}$ = Aug($X_{D}$);

            \tcp{Encoder}
        $Z_{RGB}^{inv}, Z_{RGB}^{spc}  \leftarrow Enc_{RGB}(X_{RGB})$\;
        $Z_{D}^{inv}, Z_{D}^{spc}  \leftarrow Enc_{D}(X_{D})$\;
    
        \For{$m \in \{RGB, D\}$}{
            \tcp{Define complementary modality}
            $\overline{m}\!=\!D$ \textbf{ if } $~m\!=\!RGB~$  \textbf{ else } $\overline{m}\!=\!RGB$

            \tcp{Feature mixup} 
            $\tilde{Z}_{m}^{inv} \leftarrow \lambda Z_{\overline{m}}^{inv} + (1-\lambda)Z_{m}^{inv}$\;

            \tcp{Main Semantic Segmentation task}
            $S_{m} \leftarrow Dec_m([Z_{m}^{inv}:Z_{m}^{spc}])$\;
            $\tilde{S}_{m} \leftarrow Dec_m([\tilde{Z}_{m}^{inv}:\tilde{Z}_{m}^{spc}])$\;
            Compute $\mathcal{L}^m_{seg}$ using ($S_{m}, \tilde{S}_{m}, Y$) with eq.~\eqref{eq:seg}\;

            \tcp{Auxiliary Semantic Segmentation task}
            $A_{m}^{inv} \leftarrow Dec_{Aux}(Z_{m}^{inv})$\;
            $A_{m}^{spc} \leftarrow Dec_{Aux}(Z_{m}^{spc})$\;
            Compute $\mathcal{L}^m_{aux}$ using ($ A_{m}^{inv}, A_{m}^{spc}, Y$) with eq.~\eqref{eq:aux}\;

            \tcp{Disentanglement contrainsts}
            Compute $\mathcal{L}^m_{\perp}$ using ($ Z_{m}^{inv},  Z_{m}^{spc}$) with eq.~\eqref{eq:perp}\;
            Compute $\mathcal{L}^m_{con}$ using  ($Z_{m}^{inv}, Z_{\overline{m}}^{inv}$)  with eqs.~\eqref{eq:pooling}-\eqref{eq:con}\; 
            }
        $\mathcal{L}_{tot} = \sum_{m \in\{RGB, D\}} \mathcal{L}^m_{seg} + \mathcal{L}^m_{\perp} +  \mathcal{L}^m_{con} + \mathcal{L}^m_{aux}$
    
            Update weights of $(Enc_{RGB}, Enc_{D}, Dec_{RGB}, Dec_D, Dec_{Aux})$ by back-propagating $\mathcal{L}_{tot}$
        }
    }
    \Return $(Enc_{RGB}, Enc_{D}, Dec_{RGB}, Dec_D)$
\end{algorithm}

\section{Experiment}
\label{sec:expe}
To assess the behavior of our framework, \method{}, we conducted a comprehensive experimental evaluation using three RGBD benchmarks, comparing our approach against recent competitors in Cross-Modal Knowledge Distillation for semantic segmentation. Furthermore, we performed an ablation study to examine the contributions of individual \method{} components and a sensitivity analysis on the hyperparameter $\lambda$, which controls the feature mixup strategy across modalities. Finally, we analyze and discuss the computational requirements of competing methods in terms of both training time and model size (parameter counts), emphasizing the advantages provided by \method{} over competitors based on the conventional teacher-student paradigm.

\paragraph{\textbf{Benchmarks:}} We selected three RGBD semantic segmentation datasets spanning diverse domains to ensure a broad evaluation: indoor scene segmentation, aerial imagery and synthetic drone flight data. Specifically, we considered the following benchmarks:
\begin{itemize}
    \item \textbf{NYU Depth v2} \cite{nyudepth}: dataset consisting of 1,449 pairs of indoor RGB and Depth images, labeled with 40 semantic classes. Each image has a resolution of $480 \times 640$ and is captured using Microsoft’s Kinect. Following prior works \cite{acnet,Girdhar2022OmnivoreAS, sigma}, we split the dataset into 795 training pairs and 654 test pairs;
    \item \textbf{Potsdam}~\cite{potsdam}: a remote sensing dataset comprising 38 scenes of true orthophotos with a ground resolution of 5 cm, annotated with 6 semantic classes. The dataset includes four-channel visual images (R-G-B-IR) and corresponding Digital Surface Models (DSM). For our experiments, we used IR-G-B images and the provided normalized DSMs. Each high-resolution 6,000 $\times$ 6,000 scene was divided into $500 \times 500$ crops with stride 1 and further resized to $256 \times 256$ due to computational constraints. This resulted in a total of 5,472 images. We followed the same training/test split as described in \cite{Kieu2023GeneralPurposeMT};
    \item \textbf{Mid-Air} \cite{midair}: a synthetically generated dataset designed for low-altitude drone flight segmentation, containing 79 minutes of flight data across different weather and seasonal conditions. It includes RGB images and stereo disparity depth maps annotated with 13 semantic classes. Given the large dataset size (over 400k frames) and computational limitations, we selected only a subset of images generated using Unreal Engine’s PLE plugin during the spring season. We further subsampled the dataset by selecting one frame every 8 and downscaling the resolution from 1,024 $\times$ 1,024 to $256 \times 256$. This resulted in 6,859 images, which were split into training and test sets following the original benchmark.
\end{itemize}

\paragraph{\textbf{Competing methods:}}
We compare our approach against several baselines and state-of-the-art CMKD methods for semantic segmentation. Specifically, we evaluate:
\begin{itemize}
	\item Single-modality, either RGB or Depth, models which do not receive any distillation supervision (referred to as single-modality);
	\item A full multimodal architecture, corresponding to the teacher model (referred to as \textit{multimodal});
	\item Two standard knowledge distillation (KD) baselines \cite{Xue2022TheMF} (referred to as $KDv1$ and $KDv2$);
	\item Four state-of-the-art CMKD frameworks, especially tailored for semantic segmentation in multi-modal scenario.
\end{itemize}

For KD baselines, we adopt the approaches proposed in \cite{Xue2022TheMF}. These follow a standard KD framework (Equation \ref{eq:kd}), where $\alpha$ controls the balance between task-specific loss and knowledge distillation. In particular, $KDv1$ sets $\alpha = 0$, meaning the student model learns exclusively from the teacher’s soft labels, whereas $KDv2$ uses $\alpha = 0.5$, combining both the original ground-truth labels and the teacher’s soft labels equally.

Regarding state-of-the-art CMKD competing frameworks, we consider the following approaches from the recent literature:
\begin{itemize}
    \item \textbf{KD-Net} \cite{minhao}: originally designed for medical imaging, KD-Net transfers knowledge from a multimodal teacher network to a single-modal student to handle missing modalities. 
    It employs a generalized KD framework \cite{LopezPaz2015UnifyingDA}, utilizing both the teacher’s soft labels and bottleneck logits alongside a task-specific loss (binary cross-entropy and Dice loss). 
    \item Masked Generative Distillation (\textbf{Masked Dist.}) \cite{masked}: introduces a generative distillation task where the student learns to reconstruct a corrupted feature map using the teacher’s features as a reference. The final loss consists of a task-specific segmentation loss and a generative distillation term. For the experimental evaluation, we use the encoder’s output as feature map to reconstruct.
    \item \textbf{ProtoKD}~\cite{prototype}: combines prototype learning with traditional knowledge distillation and segmentation loss. This method captures semantic correlations across the entire dataset by modeling intra- and inter-class feature variations, transferring similarity maps from the teacher to the student. For the experimental evaluation, we consider the features of the decoder just before the logits computation.
    \item Layer-wise Angular Distillation (\textbf{LAD})~\cite{liu} and  Channel-wise Angular Distillation (\textbf{CAD})~\cite{liu}: these methods extend conventional KD approaches by incorporating angular constraints on features. Similar to KD-Net, they perform distillation on both bottleneck features and logits. However, LAD applies layer-wise angular constraints, while CAD operates on channel-wise angular representations. 
\end{itemize}

\paragraph{\textbf{Experimental Settings:}}
We adopted a consistent training setup across all methods and experiments, with 140 training epochs and batch size of 8. Optimization was performed using the AdamW optimizer with a staring learning rate of $10^{-8}$ and a learning rate schedule with 10 linear warmup epochs, reaching a target learning rate of $10^{-4}$, followed by polynomial decay with power 0.9. Regarding \method{}, the temperature $\tau$ and the feature mixup $\lambda$ hyperparameters were set to 0.07 and 0.35, respectively. For data augmentation, we use random flipping, scaling, and cropping for both RGB and Depth images and color jittering only for RGB images, following common practices from previous research~\cite{acnet, sigma}.
Model performance on the test set has been evaluated using the mean Intersection over Union (mIoU) metric. All experiments were conducted on a single NVIDIA A40 GPU with 48 GB of memory.

\paragraph{\textbf{Implementation details:}}
To ensure a fair comparison, all the competing approaches share the same architecture, which follows a convolutional encoder-decoder design. For each modality, the encoder is based on a ResNet-50 network with dilated convolutions\footnote{The final pooling operation is removed and replaced with a $conv_{1 \times 1}$ projection, followed by batch normalization and a ReLU activation, reducing the feature dimensionality from 2048 to 1024.} and initialized with ImageNet-pretrained weights. In the Depth single-modality model, the first-layer weights are initialized by averaging the three-channel pretrained weights into a single-channel representation.

Segmentation outputs are generated using the DeepLabV3+ model~\cite{deeplabv3+}, which integrates Atrous Spatial Pyramid Pooling (ASPP) and a skip connection linking the second convolutional layer of the ResNet backbone to the decoder. 

\begin{figure}[!ht]
    \centering
    \includegraphics[width=1\textwidth]{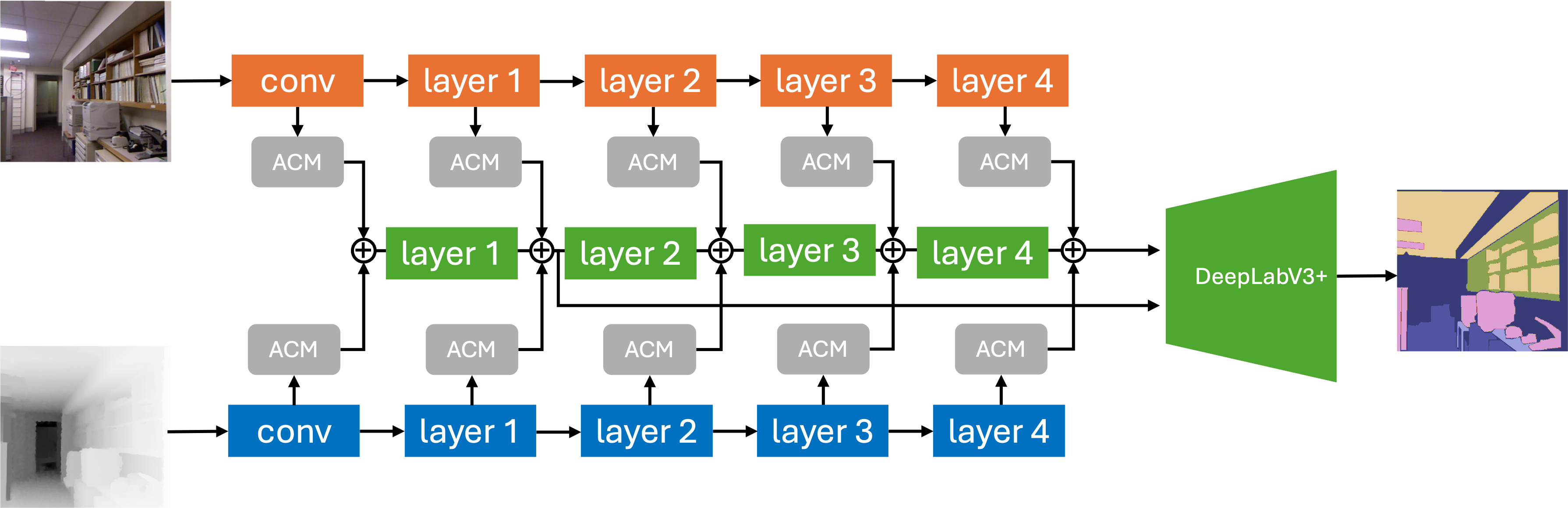}
    \caption{Teacher model architecture used for the competing methods. It consists of ResNet50 branches for i) RGB ii) Depth and iii) fused representation encoding with an Attention Complementary Module (ACM) as proposed in ACNet~\cite{acnet}. A DeepLabV3+ decoder is added to generate semantic segmentation predictions.  \label{fig:teacher}}
\end{figure}

The teacher model follows the ACNet~\cite{acnet} architecture, a commonly adopted multi-modal semantic segmentation framework for RGBD data. It consists of two ResNet-50 branches dedicated to RGB and Depth modalities, alongside a third ResNet-50 branch for fusing per-modality features. The fusion process is further refined through an Attention Complementary Module (ACM), which applies attention pooling, a $1 \times 1$ convolution followed by a sigmoid activation function, as introduced in ACNet. The DeepLabV3+ decoder then processes the fused representation. Figure~\ref{fig:teacher} provides an overview of the teacher model architecture.

\subsection{Results}
We present in Table~\ref{tab:main} a comparison between the performances achieved by \method{} and the competing methods described in the previous section, in terms of mIoU score. Specifically, we include the multi-modal teacher, single-modality models, standard KD baselines (Equation \ref{eq:kd}) and state-of-the-art competitors. We highlight models that outperform the single-modality baseline with a green arrow and those that underperform the same baseline with a red arrow. To ensure a comprehensive evaluation, we assess each benchmark in two cross-modal distillation scenarios, transferring knowledge from multi-modal RGBD data to either the RGB or Depth modality.

We observe that \method{} consistently outperforms all competitors across all benchmarks in both RGB and Depth cross-modal distillation scenarios. For the NYUDepth and Mid-Air datasets, the single-modality models are  outperformed by the multimodal teacher. However, in the Postdam benchmark, both \method{} and Masked Dist. produce RGB-based models that surpass the multimodal approach, achieving mIoU scores of 76.13 and 76.09, respectively, compared to the 74.98 mIoU achieved by the multimodal teacher. Notably, our framework stands out as the only one that consistently demonstrates improvements (green arrows) over the single-modality baselines across all cross-modal scenarios, delivering results that surpass the state-of-the-art methods in Cross-Modal Knowledge Distillation for the considered RGBD benchmarks.

\begin{table*}[!ht]
    \centering
    \normalsize
\begin{tabular}{l|ll|ll|ll}
\multicolumn{1}{c|}{\multirow{2}{*}{\textbf{Model}}} & \multicolumn{2}{c|}{\textbf{NYUDepth}}                            & \multicolumn{2}{c|}{\textbf{Potsdam}} & \multicolumn{2}{c}{\textbf{Mid-Air}} \\ 
\multicolumn{1}{c|}{}                       & \multicolumn{1}{l}{RGB} & \multicolumn{1}{l|}{Depth} & \multicolumn{1}{l}{RGB}        & \multicolumn{1}{l|}{Depth}       & \multicolumn{1}{l}{RGB}        & \multicolumn{1}{l}{Depth}  \\ \hline 
multimodal & \multicolumn{2}{c|}{46.92} & \multicolumn{2}{c|}{74.98} & \multicolumn{2}{c}{51.21} \\ 
single-modality & 42.64   & 36.01 & 75.73  & 42.47 & 47.84  &  47.07 \\ \hline \hline 
KDv1~\cite{XueGRZ23} & 43.43 \greenArrow  & 36.44 \greenArrow & 66.32 \redArrow  & 39.20 \redArrow & 47.36 \redArrow  & 45.80 \redArrow \\ 
KDv2~\cite{XueGRZ23} & 43.86 \greenArrow & 36.91 \greenArrow & 66.24 \redArrow & 39.38 \redArrow & 47.62 \redArrow & 45.88 \redArrow \\ \hline 
KD-Net~\cite{minhao} & 42.78 \greenArrow & 36.36 \greenArrow & 73.82 \redArrow & 41.85 \redArrow & 48.32 \greenArrow & 46.22 \redArrow \\ 
Masked Dist.~\cite{masked} & 40.97 \redArrow & 34.93 \redArrow & 76.09 \greenArrow & 42.43 \redArrow & 47.60 \redArrow & 47.40 \greenArrow \\ 
ProtoKD~\cite{prototype} & 43.82 \greenArrow & 37.28 \greenArrow & 66.64 \redArrow & 39.27 \redArrow & 47.11 \redArrow & 45.45 \redArrow \\
LAD~\cite{liu} & 43.62 \greenArrow & 36.86  \greenArrow & 66.80 \redArrow & 39.31 \redArrow & 48.01 \greenArrow & 46.98 \redArrow \\ 
CAD~\cite{liu} & 43.48 \greenArrow & 37.16 \greenArrow &  66.43 \redArrow & 38.89 \redArrow & 48.21 \greenArrow & 47.09 \greenArrow \\  \hline 
\method{} & \textbf{44.85} \greenArrow & \textbf{37.60} \greenArrow & \textbf{76.13} \greenArrow & \textbf{42.78} \greenArrow & \textbf{48.37} \greenArrow & \textbf{47.91} \greenArrow\\ 
\end{tabular}
\caption{Mean Intersection over Union (mIoU) performances over the three considered benchmarks, comparing our model with the multi-modal teacher and single-modality models, as well as state-of-the-art competitors for CMKD semantic segmentation; green and red arrows indicate, respectively, improvement or reduction of scores with respect to the single-modality model. }
\label{tab:main}
\end{table*}


\paragraph{\textbf{Ablation}}
Table~\ref{table:abla} presents the results of our ablation study, examining the contribution of individual components and loss terms in \method{}. Our analysis reveals that the most significant performance drops occur when removing the auxiliary loss ($\mathcal{L}{aux}$) and the contrastive loss ($\mathcal{L}{con}$), indicating their crucial role in the framework. The impact of other components and loss terms remains comparable, with variations depending on the dataset. Overall, the highest performance is consistently achieved when all components are included, highlighting the rationale behind \method{}. 

\begin{table*}[!ht]
    \centering
    \normalsize
\begin{tabular}{l|ll|ll|ll|l}
& \multicolumn{2}{c|}{\textbf{NYUDepth}}                            & \multicolumn{2}{c|}{\textbf{Potsdam}} & \multicolumn{2}{c}{\textbf{Mid-Air}} & \textbf{Avg} \\
\multicolumn{1}{c|}{}                       & \multicolumn{1}{c}{RGB} & \multicolumn{1}{c|}{Depth} & RGB        & Depth       & RGB        & Depth & \\ \hline
w/o $\mathcal{L}_{\perp}$ & 43.98 & 37.24 & 75.88 & 42.48 & 48.15 & 47.58 & 49.22\\ \hline 
w/o $\mathcal{L}_{con}$ & 43.10 & \textbf{37.96} & 75.53 & 42.31 & 48.46 & 47.26 & 49.11\\ \hline 
w/o $\mathcal{L}_{aux}$ & 44.84 & 37.62 & 75.55 & \textbf{42.99} & 47.08 & 46.44 & 49.09\\ \hline 
w/o \textit{mixup} & 44.82 & 37.48 & 75.52 & 42.36 & 48.19 & 47.47 & 49.31\\ \hline 
w/o \textit{dec. aug.} &  43.62 & 37.49 & 75.92 & 42.37 & 48.31 & 47.30 & 49.17\\ \hline \hline 
Original & \textbf{44.85} & 37.60 & \textbf{76.13} & 42.78 & \textbf{48.37} & \textbf{47.91} & \textbf{49.61}\\ \hline 
\end{tabular}
\caption{Analysis of the contributions of all the
components of \method{} in terms of mIoU. \label{table:abla}}
\end{table*}

\paragraph{\textbf{Sensitivity Analysis}}
We explored the impact of varying the mixup hyperparameter $\lambda$ (Equation \ref{eq:mixup}) from 0.05 to 0.5, adjusting the degree of feature mixup between domain-invariant RGB and Depth features. As shown in Table \ref{tab:mixup}, performance remains relatively stable across this range, with no significant variation as highlighted by the standard deviation. 

\begin{figure*}[h]
    \centering
    \setlength\tabcolsep{-10pt}  
    \renewcommand{\arraystretch}{0.1} 
    \centering
    \begin{tabular}{cccccc}
     \includegraphics[width=0.22\textwidth]{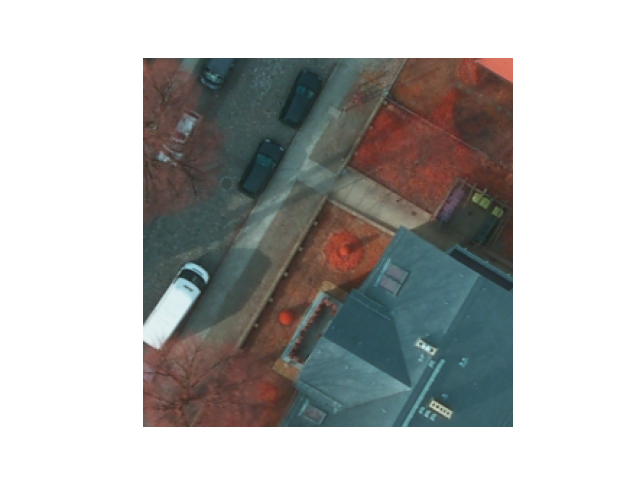} &
     \includegraphics[width=0.22\textwidth]{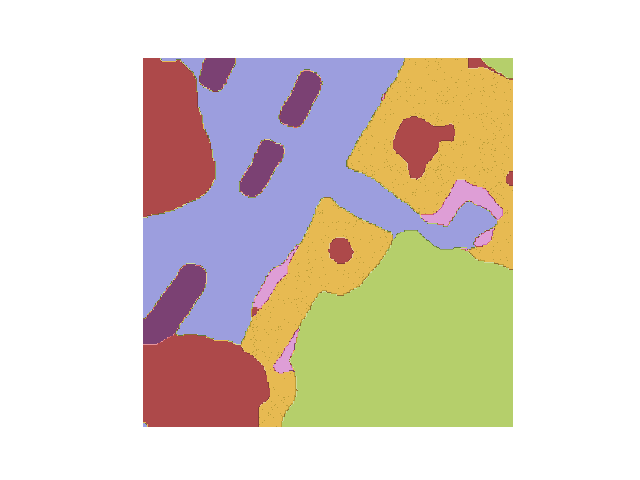} &
     \includegraphics[width=0.22\textwidth]{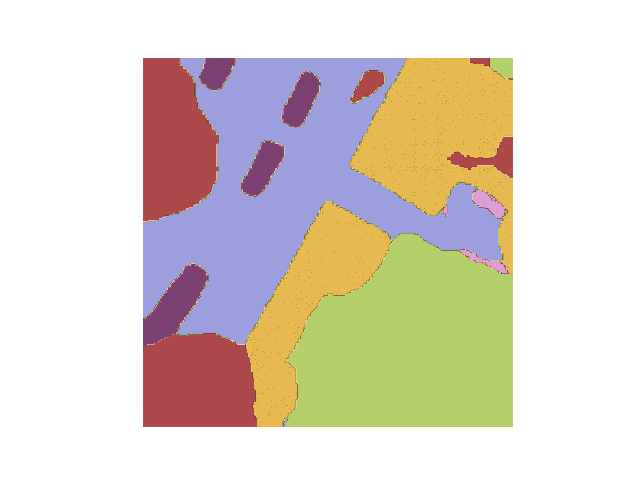} &
     \includegraphics[width=0.22\textwidth]{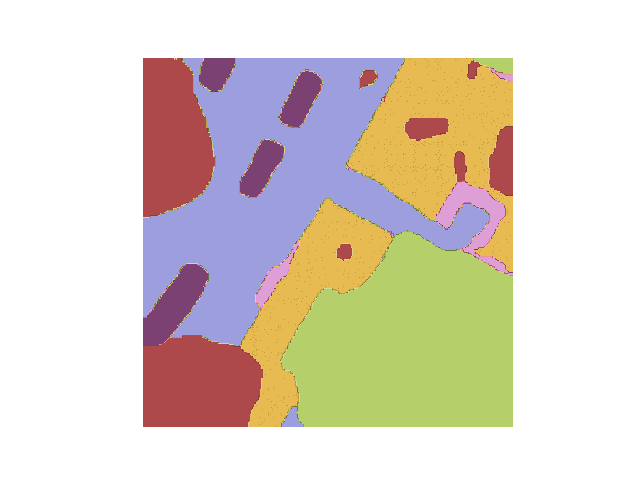} &
     \includegraphics[width=0.22\textwidth]{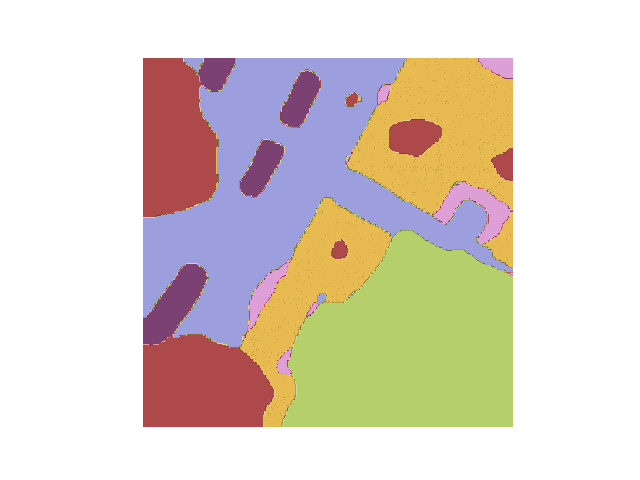} &
     \includegraphics[width=0.22\textwidth]{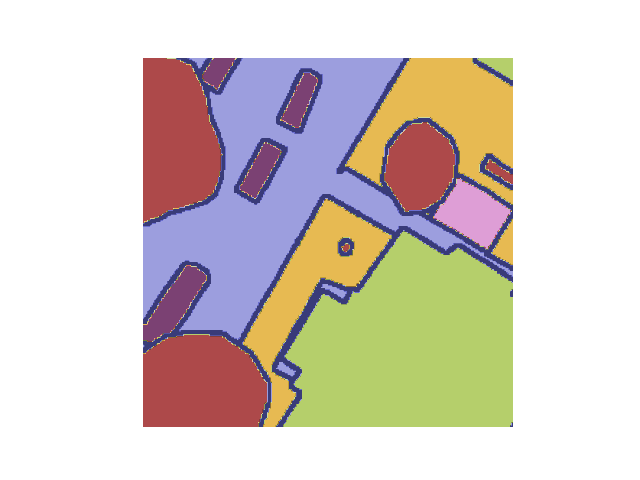}  \\
     \includegraphics[width=0.22\textwidth]{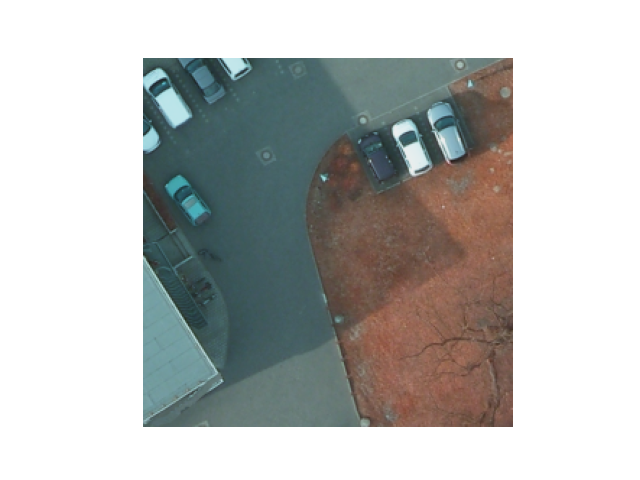} &
     \includegraphics[width=0.22\textwidth]{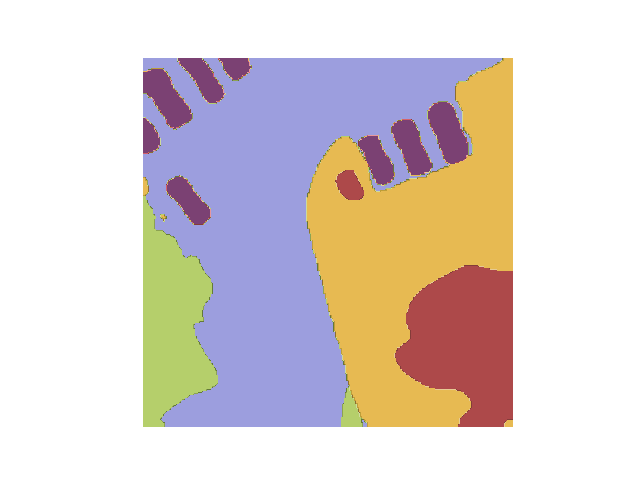} &
     \includegraphics[width=0.22\textwidth]{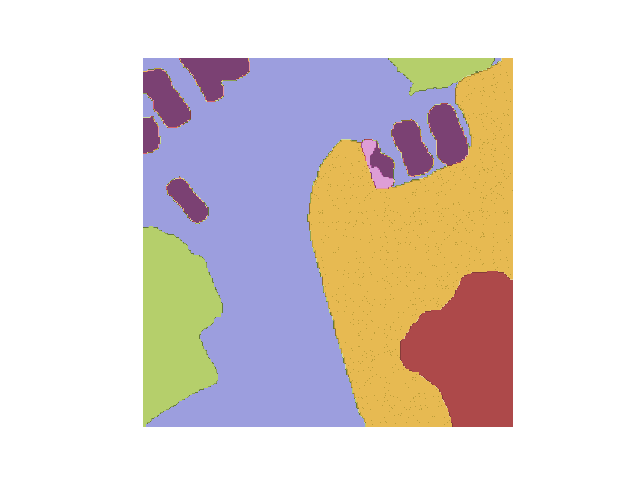} &
     \includegraphics[width=0.22\textwidth]{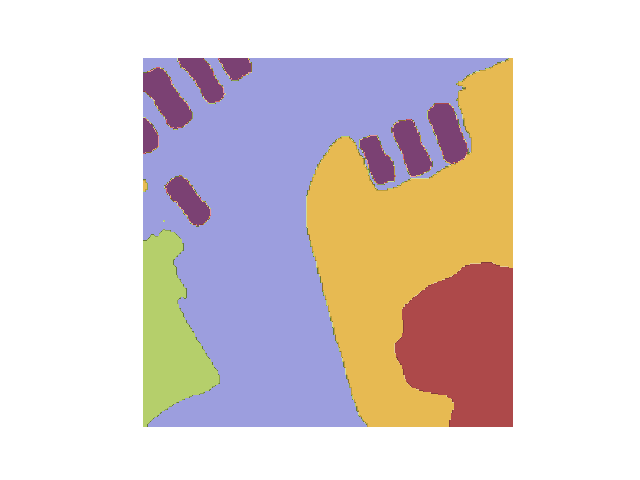} &
     \includegraphics[width=0.22\textwidth]{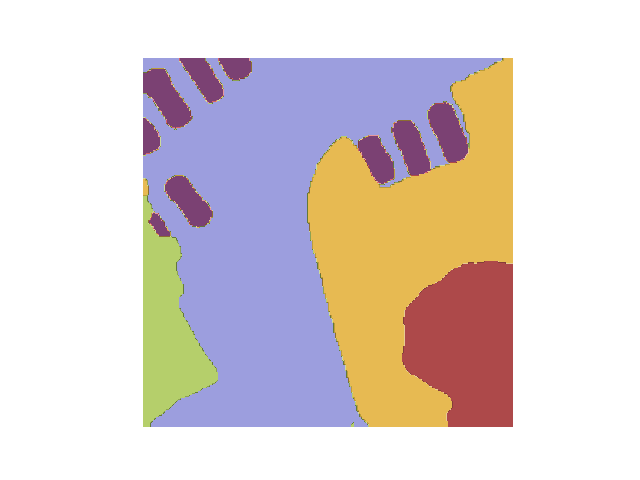} &
     \includegraphics[width=0.22\textwidth]{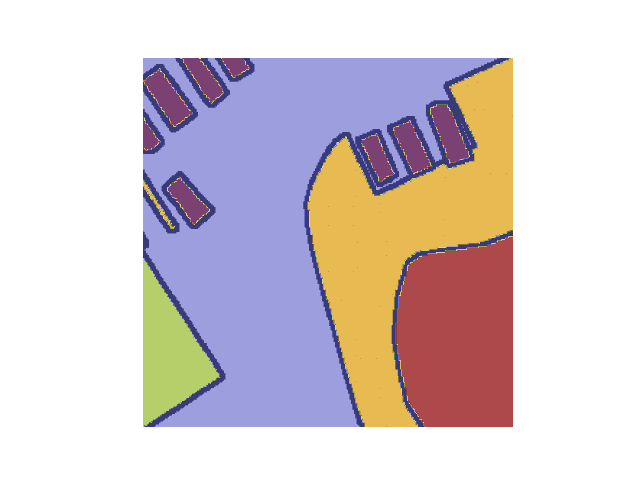}  \\
     \includegraphics[width=0.22\textwidth]{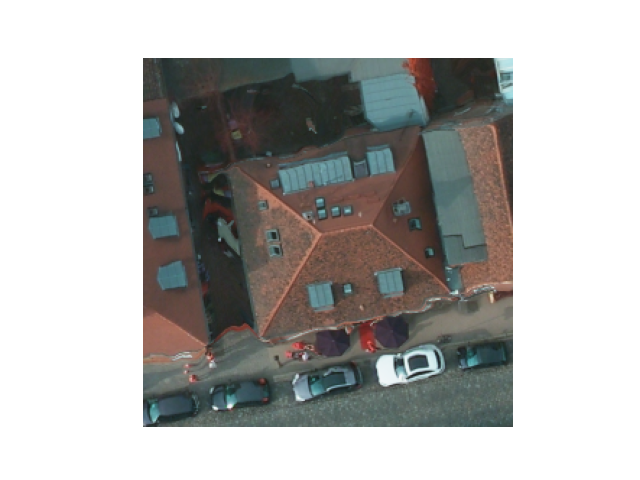} &
     \includegraphics[width=0.22\textwidth]{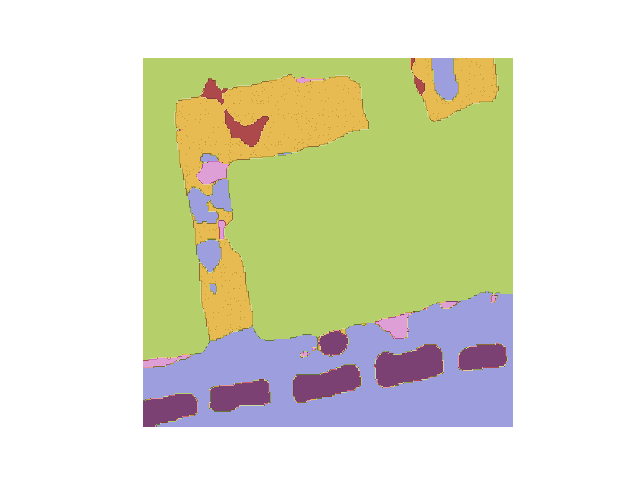} &
     \includegraphics[width=0.22\textwidth]{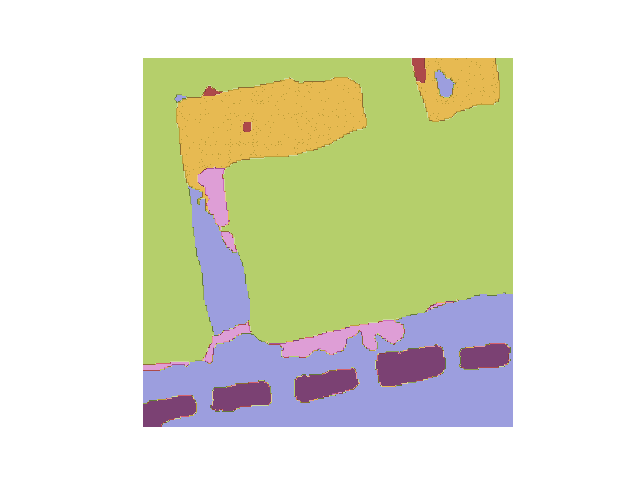} &
     \includegraphics[width=0.22\textwidth]{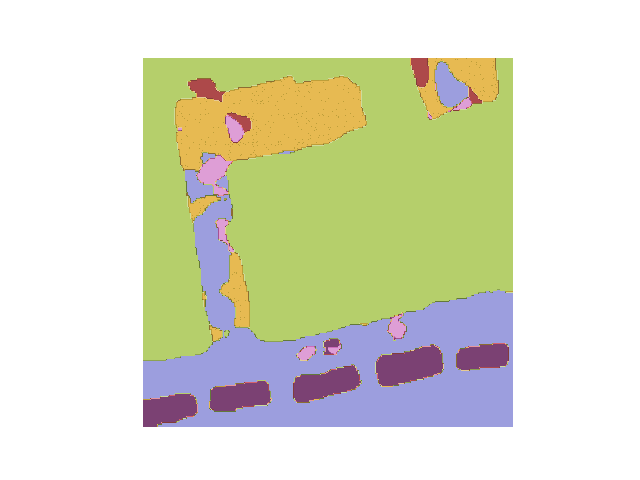} &
     \includegraphics[width=0.22\textwidth]{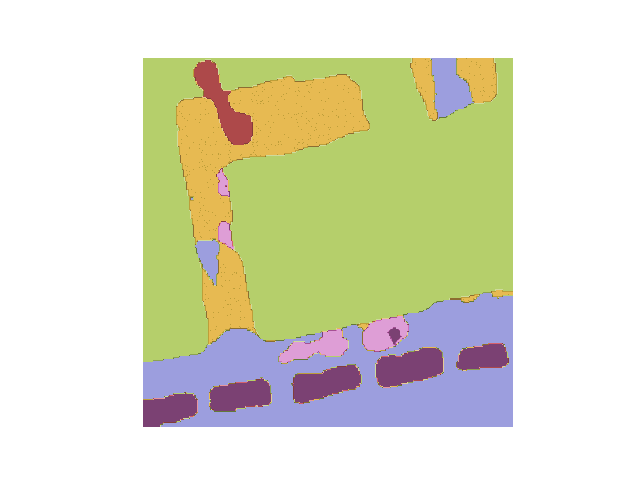} &
     \includegraphics[width=0.22\textwidth]{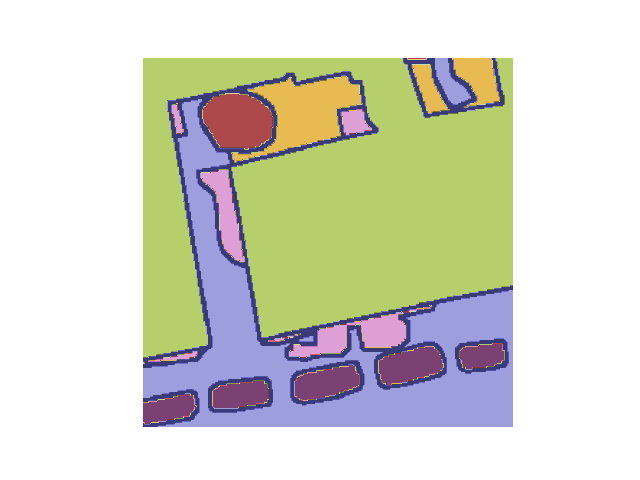} \\
     RGB & single-modality & KD-Net & Masked Dist. & \method & Ground Truth
    \end{tabular}
    \caption{Example of qualitative results from Potsdam dataset.\label{fig:potsdam}}
\end{figure*}

\paragraph{\textbf{Segmentation examples}}
Some qualitative segmentation examples on the Potsdam dataset are presented in Figure~\ref{fig:potsdam}. Here, we compare our method with best performing competitors (KD-Net and Masked Dist.) and the single-modality baseline. The analysis focuses on the RGB modality, as it provides greater visual detail. The results clearly show that all the CMKD frameworks provide a more precise and reliable segmentation mask compared to the one produced by the single-modality baseline. Among the different approaches, we can observe that the quality of segmentation examples is consistent with the quantitative results we have reported above.

\begin{table}[h]
\centering
\begin{tabular}{l|rr|rr|rr}
\multirow{2}{*}{\textbf{$\lambda$}}  & \multicolumn{2}{c|}{\textbf{NYUDepth}}                            & \multicolumn{2}{c|}{\textbf{Potsdam}} & \multicolumn{2}{c}{\textbf{Mid-Air}}  \\
\multicolumn{1}{c|}{}                       & \multicolumn{1}{c}{RGB} & \multicolumn{1}{c|}{Depth} & RGB        & Depth       & RGB        & Depth \\ \hline
0.05 & 44.55 & 37.66 & 75.90 & 42.66 & 48.16 & 46.82 \\
0.1 & 44.67 & 37.72 & 76.09 & 42.71 & 47.95 & 46.96 \\
0.2 & 44.64 & 37.82 & 76.06 & 42.59 & 48.17 & 47.04 \\
0.35 & 44.85 & 37.60 & 76.13 & 42.78 & 48.37 & 47.91 \\
0.5 & 44.53 & 37.50 & 75.99 & 42.39 & 48.13 & 47.41 \\ \hline \hline
std & 0.13 & 0.12 & 0.09 & 0.15 & 0.15 & 0.44 \\
\end{tabular}
\caption{Sensitivity analysis on the feature mixup hyperparameter $\lambda$. \label{tab:sensi}}
\label{tab:mixup}
\end{table}

\begin{figure*}[h!]
    \centering
    \setlength\tabcolsep{4pt}  
    \renewcommand{\arraystretch}{1} 
    \centering
    \begin{tabular}{ccccc}
     \includegraphics[width=0.18\textwidth]{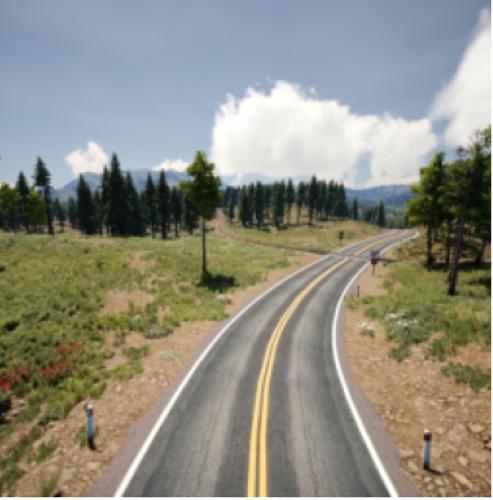} &
     \includegraphics[width=0.18\textwidth]{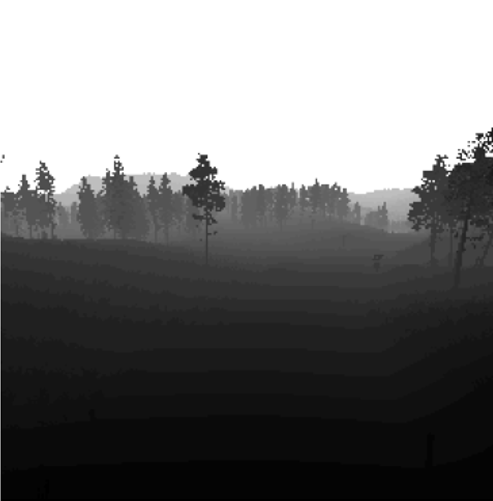} &
     \includegraphics[width=0.18\textwidth]{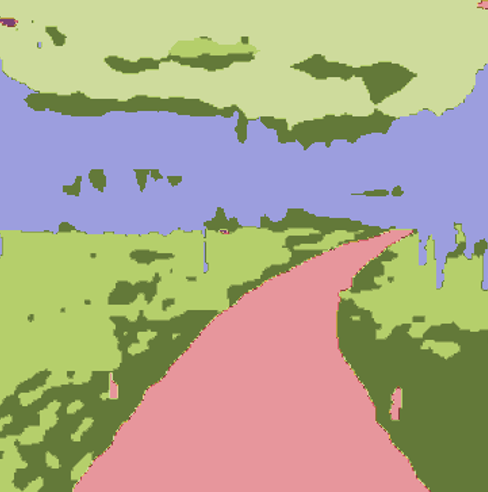} &
     \includegraphics[width=0.18\textwidth]{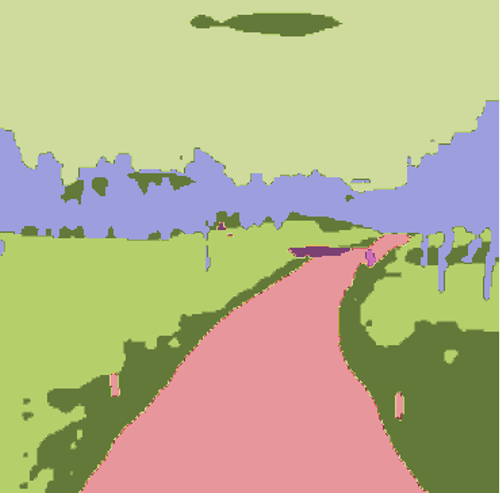} &
     \includegraphics[width=0.18\textwidth]{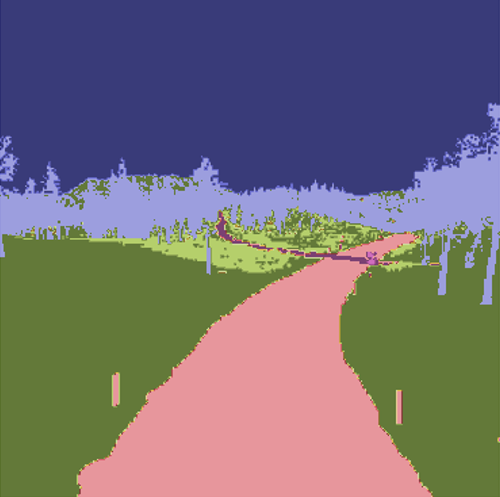} \\
     \includegraphics[width=0.18\textwidth]{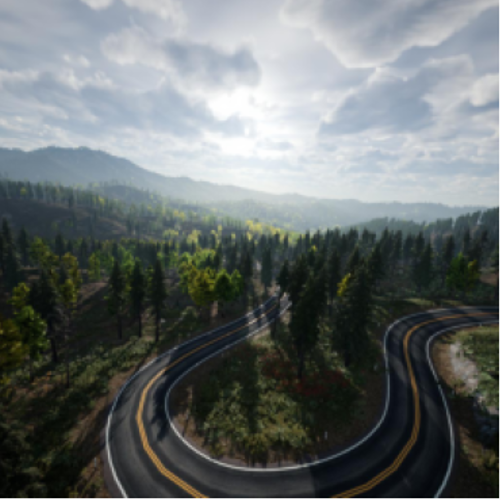} &
     \includegraphics[width=0.18\textwidth]{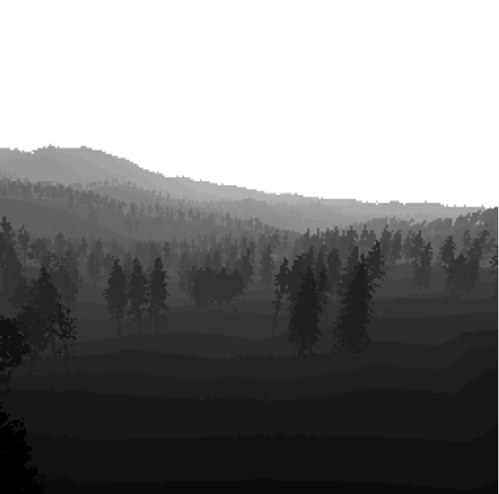} &
     \includegraphics[width=0.18\textwidth]{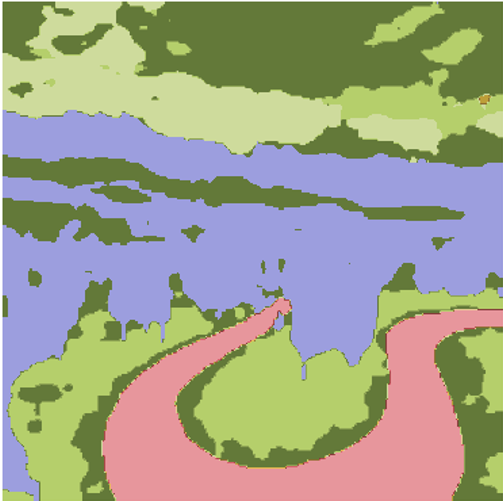} &
     \includegraphics[width=0.18\textwidth]{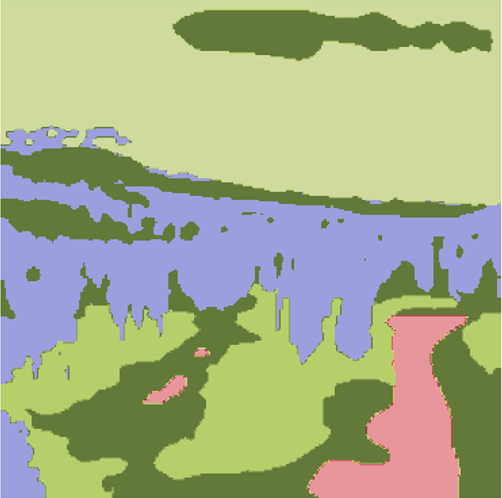} &
     \includegraphics[width=0.18\textwidth]{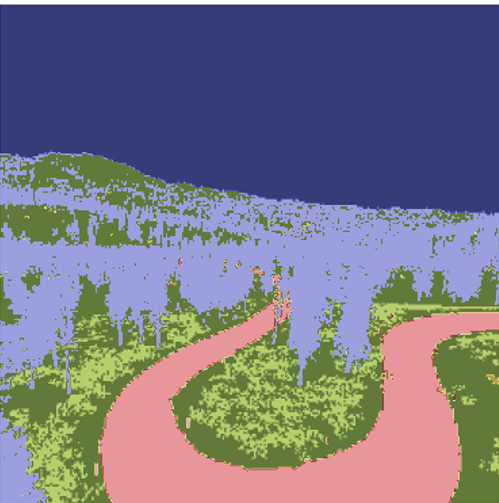} \\
     \includegraphics[width=0.18\textwidth]{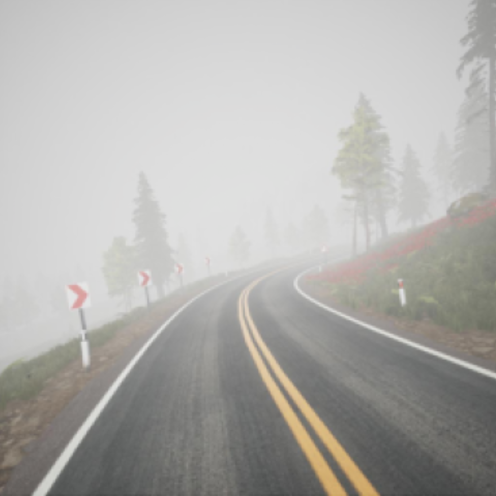} &
     \includegraphics[width=0.18\textwidth]{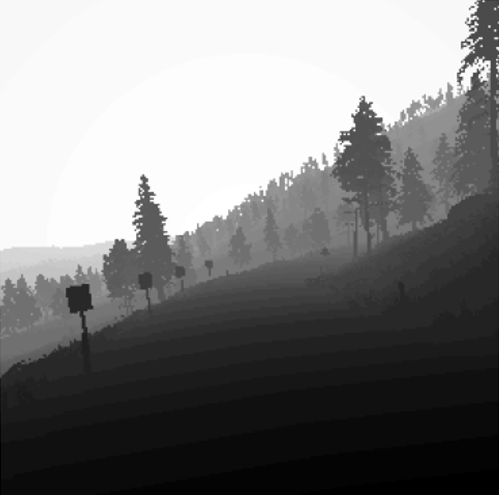} &
     \includegraphics[width=0.18\textwidth]{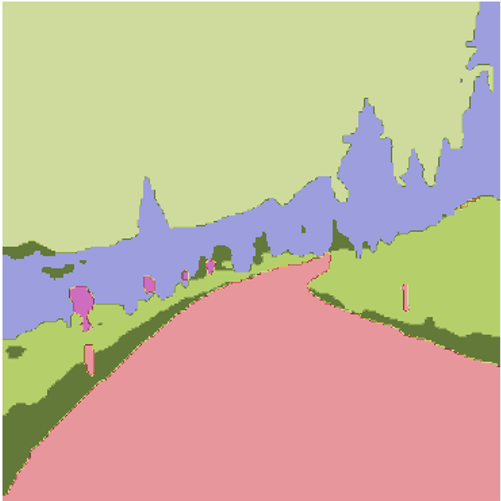} &
     \includegraphics[width=0.18\textwidth]{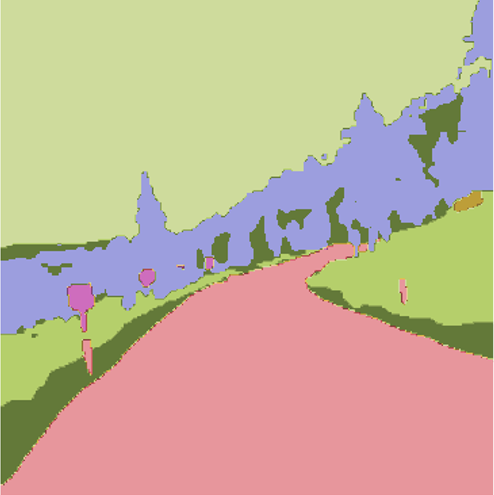} &
     \includegraphics[width=0.18\textwidth]{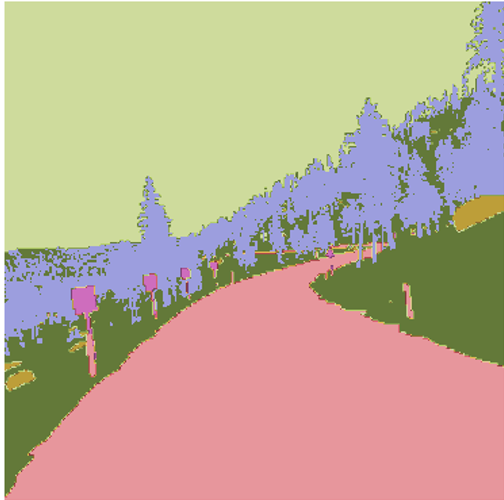} \\
     RGB & Depth & RGB branch & Depth branch & Ground Truth
    \end{tabular}
    \caption{Example of qualitative results from CroDiNo-KD predictions over the MidAir dataset.\label{fig:branches}}
\end{figure*}

To further inspect the behavior of our model, in Figure \ref{fig:branches} we depict the output of each per-modality branch, separately, on a few samples coming from the MidAir dataset. It could be noted that the input features may provide complementary information for the segmentation task, for example, in the second row the road is perfectly detected via the RGB sensor, while in the third row the Depth map provides useful information given the lack of visibility, due to fog, on the RGB image. 

\paragraph{\textbf{Training time and model size}}
To further emphasize the advantages of \method{}, we compare models performance in terms of total training time and model size (parameters count). Table \ref{tab:time} presents the complete training time for all competing methods on the MidAir dataset for training both RGB and Depth single-modality models. We report the training time\footnote{Training times are reported in GPU hours, meaning the equivalent training duration without parallelization.} 
for the distillation process (referred as \textbf{Main}), the one for the teacher training (referred as \textbf{Teacher}) and the total one (referred as \textbf{Tot.})
\method{} exhibits the shortest training time for the distillation process, completing both RGB and Depth single-modality models training in less than twenty-one hours. Furthermore, unlike our approach, all CMKD methods require pre-training a teacher model, adding an extra 14 hours overhead to the total training time. 
Such analysis clearly demonstrates the advantage, in terms of training time, of \method{} over standard teacher/student CMKD frameworks.

Table \ref{tab:size} compares the number of parameters required by competing frameworks during training. We categorize the parameters into: those required for the main architecture (\textbf{Main}), those used for auxiliary tasks which are discarded during inference such as the generative decoder in the \textit{Masked Dist.} model (\textbf{Aux}), the parameters of the teacher model (\textbf{Teacher}) and the total per framework parameters  (\textbf{Tot.}). 
\method{} has fewer parameters in its main architecture compared to competing models, due to the practical choice to reduce the encoder's extracted features to accommodate the disentanglement representation process. The auxiliary parameters in \method{}, associated with the auxiliary decoder, remain negligible compared to the overall model size. Furthermore, by eliminating the need for a computationally demanding multi-modal teacher, our approach requires less than half the parameters of the second smallest CMKD framework, thus highlighting the parameter-efficient design of \method{}.

\begin{table}[ht!]
\parbox{.45\linewidth}{
\centering
\begin{tabular}{l||c|c||c|}
\textbf{Model} & \multicolumn{3}{c}{\textbf{GPU hours}} \\ \hline 
& \textbf{Main} & \textbf{Teacher} & \textbf{Tot.} \\ \hline
\hline
Single-Modality   & 14h 52m  & - & 14h 52m\\ \hline
KDv1 / KDv2 & 22h & 14h & 36h \\ \hline
KD-Net  & 22h 46m & 14h & 36h 46m\\
Masked Dist. & 47h 22m & 14h & 61h 22m\\
ProtoKD  & 25h 02m & 14h & 39h 02m\\
LAD   & 22h 26m & 14h & 36h 26m\\
CAD  & 38h 24m  & 14h & 52h 24m\\ \hline 
CroDiNo-KD  & 20h 30m & - & 20h 30m\\
\end{tabular}
\caption{Training time in GPU hours.}
\label{tab:time}
}
\hfill
\parbox{.45\linewidth}{
\centering
\begin{tabular}{l||c|c|c||c|}
\textbf{Model} & \multicolumn{4}{c}{\textbf{Num. of Params.}} \\ \hline
& \textbf{Main} & \textbf{Aux} & \textbf{Teacher} & \textbf{Tot.}\\ \hline \hline
Single-Modality & 80M & - & - & 80M \\ \hline
KDv1 / KDv2 & 80M & - & 98M & 178M \\ \hline
KD-Net  & 80M & - & 98M & 178M \\
Masked Dist.  & 80M & 150M & 98M & 328M \\
ProtoKD  & 80M & - & 98M  & 178M \\
LAD/CAD   & 80M & - & 98M & 178M \\ \hline 
CroDiNo-KD  & 68M & 5M & - & 73M \\
\end{tabular}
\caption{Models' size in terms of parameters counts at training time.}
\label{tab:size}
}
\end{table}

\section{Conclusion}
\label{sec:conclu}
In this paper, we propose \method{}, a novel framework for RGBD Cross-Modal Knowledge Distillation (CMKD). Unlike conventional teacher/student approaches, our framework facilitates knowledge transfer between single-modality models without requiring a multi-modal teacher. This is achieved by leveraging disentanglement representation learning, contrastive learning and decoupled data augmentation. Through carefully designed loss functions, our method structures the internal manifolds of the single-modality models to account for both modality-invariant and modality-specific features. This approach harnesses the synergy between RGB and Depth modalities to enhance semantic segmentation performance in scenarios where mismatches exist between the data modalities accessible during training and inference.  Our evaluation demonstrates the quality of \method{} over baselines and state-of-the-art CMKD frameworks, considering both classification performance and computational efficiency during training. Furthermore, our findings invite reconsidering the traditional teacher/student paradigm for distilling information from multi-modal data to single-modality neural networks in the context of semantic segmentation.


\bibliographystyle{splncs04}
\bibliography{biblio}

\end{document}